\def\year{2018}\relax
\documentclass[letterpaper]{article} 
\usepackage{aaai18}  
\usepackage{times}  
\usepackage{helvet}  
\usepackage{courier}  
\usepackage{url}  
\usepackage{graphicx}  
\usepackage{latexsym}
\usepackage{mathtools}
\usepackage{amsmath}
\usepackage{amssymb}
\usepackage{bm}
\usepackage{multirow}
\usepackage{subcaption}
\usepackage{esvect}
\usepackage{color}

\title{Dual Attention Network for Product Compatibility\\and Function Satisfiability Analysis}
\author{
Hu Xu\\
Department of Computer Science\\
University of Illinois at Chicago\\Chicago, IL, USA\\
hxu48@uic.edu\\
\And
Sihong Xie\\
Department of Computer Science and Engineering\\
Lehigh University\\Bethlehem, PA, USA\\
sxie@cse.lehigh.edu
\AND
Lei Shu\\
Department of Computer Science\\
University of Illinois at Chicago\\Chicago, IL, USA\\
lshu3@uic.edu\\
\And
Philip S. Yu\\
Department of Computer Science\\
University of Illinois at Chicago\\Chicago, IL, USA\\
Institute for Data Science\\Tsinghua University\\Beijing, China\\
psyu@uic.edu
}

\begin{document}
\maketitle

\begin{abstract}
Product compatibility and their functionality are of utmost importance to customers when they purchase products, and to sellers and manufacturers when they sell products. 
Due to the huge number of products available online, it is infeasible to enumerate and test the compatibility and functionality of every product. 
In this paper, we address two closely related problems: product compatibility analysis and function satisfiability analysis, 
where the second problem is a generalization of the first problem (e.g., whether a product works with another product can be considered as a special function). 
We first identify a novel question and answering corpus that is up-to-date regarding product compatibility and functionality information~\footnote{The annotated corpus can be found at \url{https://www.cs.uic.edu/~hxu/}.}. 
To allow automatic discovery product compatibility and functionality, we then propose a deep learning model called Dual Attention Network (DAN).
Given a QA pair for a to-be-purchased product, DAN learns to 1) discover complementary products (or functions), and 2) accurately predict the actual compatibility (or satisfiability) of the discovered products (or functions). 
The challenges addressed by the model include the briefness of QAs, linguistic patterns indicating compatibility, and the appropriate fusion of questions and answers. 
We conduct experiments to quantitatively and qualitatively show that the identified products and functions have both high coverage and accuracy, compared with a wide spectrum of baselines.
\end{abstract}

\section{Introduction}
\begin{table}[t]
\footnotesize
\centering
\scalebox{0.90}{
\begin{tabular}{ l | l }
\hline
\multicolumn{2}{ l }{ Microsoft Surface Pro 4 (128 GB, 4 GB RAM, Intel Core i5) } \\
\hline
Q: & Can the M processor \textbf{handle} \underline{photoshop}? \\
A: & It \textit{does} run Photoshop \textit{very well} on our internal test unit.\\
\hline
Q: & Does the surface pro 4 \textbf{support} the \underline{Google Play app store}? \\
A: & \textit{No}, it does not support Google Play. \\
\hline
Q: & can this \textbf{run} \underline{fallout 4} \\
A: & 
\begin{tabular}[t]{@{}l@{}} 
It \textit{struggles} graphically with Fallout 4 on \\low graphics settings. \\
\end{tabular} \\
\hline
Q: & Does this \textbf{connect to} a \underline{5G home wireless network}? \\
A: & 
\begin{tabular}[t]{@{}l@{}} 
We connect to a Comcast high speed wireless router for \\
Internet access and it seems to work well.\\
I \textit{can not speak} \textit{to the specific network}\\
in the question. \\
\end{tabular} \\
\hline
Q: & Can you \textbf{use} this \textbf{for} \underline{sketching}? \\
A: & 
\begin{tabular}[t]{@{}l@{}} 
To some extent .\\
It \textit{depends upon the degree to} which you intend to create . \\
\end{tabular} \\
\hline
\end{tabular}

}
\caption{An example of QA pairs for a tablet PC: function targets (the first 4 are complementary entities) are underlined in questions; function words are bolded; keywords that indicating compatibility or satisfiability are italicized.}
\label{table:sample}
\end{table}

Learning about the compatibility of a product that is functionally complementary to a to-be-purchased product
is an important task in e-commerce.
For customers, before they purchase a product (e.g., a mouse), it is natural for them to ask whether the to-be-purchased one can work properly with the intended complementary product (e.g., a laptop).
Such a query is driven by customers' needs on product functionality, where compatibility can be viewed as a special group of functions.
In fact, a function need is the very first step of purchase decision process. 
Whether a product can satisfy some function needs or not (a.k.a. satisfiability of a function need) even leads to the definition of \textit{product}.
In marketing, \textit{product} is defined as ``anything that can be offered to a market for attention, acquisition, use or consumption that might satisfy a want or need'' \cite{kotler2010principles}. 
For sellers or manufacturers, satisfiability on function needs are equally important as being fully-aware of existing and missing functions is crucial in increasing sales and improve products. 
Therefore, exchanging the information about functions is important to customers, sellers, and manufacturers.

Given its importance, however, such function (we omit ``need'' for simplicity) information is not fully available in product descriptions.
Just imagine the cost of compatibility test over the huge number of products, or sellers' intention of hiding missing functions.
Fortunately,
customers may occasionally exchange such knowledge online with other customers or sellers.
This allows us to adopt an NLP-based approach to automatically sense and harvest this knowledge on a large scale.

In this paper, we address two (2) closely related novel NLP tasks: \textit{product compatibility analysis} and \textit{product function satisfiability analysis}. They are defined as following.\\
\textbf{Product Compatibility Analysis}:
Given a corpus of texts, identify all tuples $(p_1$, $p_2$, $c)$,
where $p_1$ and $p_2$ are a pair of complementary products (entities), and $c\in\{1,2,3\}$ indicates
whether the two entities are compatible (1), not (2) or uncertain (3).\\
Note that two \emph{complementary entities} can be \emph{incompatible}.
For example, a mouse is a product functionally complementary to ``Microsoft Surface Pro 4'', as Surface Pro 4 doesn't
come with a mouse.
However, due to different interfaces, not all mouse models can work with Surface Pro 4 properly.
By slightly extending $p_2$ to a function expression (e.g., identifying ``work with Microsoft Surface Pro 4'' instead of ``Microsoft Surface Pro 4''), we have a more general task.\\
\textbf{Product Function Satisfiability Analysis}:
Given the same corpus, identify all tuples $(p_1$, $f$, $s)$,
where $f$ is a function expression, and $s\in\{1,2,3\}$ indicates
whether $p_1$ can satisfy (1) the function $f$ or not (2), or uncertain (3).\\
Note that functions derived from complementary entites are just one type of function called \textit{extrinsic function}s.
Functions in $f$ may also include functions derived from the product $p_1$ itself as \textit{intrinsic function}s.
For example, ``draw a picture'' is an intrinsic function for ``Microsoft Surface Pro 4''.
A function expression may consist a \textit{function word} (e.g., ``work with'' or ``draw'') and a \textit{function target} (e.g., ``Microsoft Surface Pro 4'' or ``picture'').
So function target is a general term of complementary entity.

Two challenges come immediately after formalizing these two tasks.
First, the quality of the tuples depends on the data source (namely, the corpus).
Corpora that are dense and accurate regarding compatibility and functionality information are preferred.
Second, although general information extraction models are available, novel models that can jointly identify the entities (or function expressions) and compatibility (or satisfiability) in an end-to-end manner are preferred.

We address the first challenge by annotating a high-quality corpus.
In particular, Amazon.com allows potential consumers to communicate with existing product owners or sellers via Product Community Question Answering (PCQA).
As an example, in Table \ref{table:sample} we show 5 QA pairs addressing the functionality of Microsoft Surface Pro 4 ($p_1$).
We can see that the first 4 questions address the extrinsic functions (on complementary entities) and the last one is an intrinsic function.
Specifically, ``photoshop'' is a \emph{compatible entity}, ``Google Play app store'' and ``fallout 4'' are \emph{incompatible entities}, ``5G home wireless network'' is a complementary entity with \emph{uncertain} compatibility, and ``use for sketching'' is an \emph{uncertain} function.
Observe that the to-be-purchased product $p_1$ can be identified using the product title of the product page.
We focus on extracting $p_2$ or $f$ from the question and detect $c$ or $s$ from the answers.
We leave infrequently-asked open questions (e.g., ``what products can this tablet work with?'') to future work and only focus on yes/no questions. 
The details of the annotated corpus can be found in Section Experimental Result.

Given the corpus, we address the second challenge by formulating these two (2) tasks as sequence labeling problems, which fuse the information from both the questions and the answers.
We propose a model called Dual Attention Network (DAN) to solve the sequence labeling problems.
DAN addresses two technical challenges.
First, the questions and answers are usually brief~\footnote{the longest question has only 82 words} with rather limited context.
Second, the polarity information in many answers are implicit (without an explicit ``Yes'' or ``No'' in the very beginning, e.g., the 1st and 3rd answers in Table \ref{table:sample}).
DAN resolves these challenges by taking the question and the answer together as a QA story (or document) and perform two reading comprehensions \cite{richardson-burges-renshaw:2013:EMNLP,rajpurkar-EtAl:2016:EMNLP2016} on such a story as side information for sequence labeling. 
For example, it may not be obvious that ``photoshop'' is a complementary entity by reading the question only.
However, the word ``run'' in the answer is a strong indicator that ``photoshop'' is a complementary entity and the word ``well'' indicates a positive polarity.
We conduct experiments quantitatively and qualitatively to show the performance of DAN.
The proposed dual attention architecture is not limited to the proposed tasks, but may potentially be applied to other QA tasks. 

\section{Related Work}
\label{sec:rw}
Complementary products are studied in the context of recommender system \cite{McAPanLes15}. 
In~\cite{McAPanLes15}, topic models are used to predict substitutes and complements (without compatibility) of a product. 
But their work takes the outputs of Amazon's recommender system as the ground truths for complements, which can be inaccurate.
Instead, we take an information extraction approach similar to the Complementary Entity Recognition (CER) task in~\cite{xu2016CER,xu2017supervised}. 
But we perform in a supervised setting on an annotated QA corpus \cite{McAYan16,xu2016pcqa}. 
We further generalize to a more fundamental task: function satisfiability analysis.
Although \cite{xu2017Fun} has a preliminary study on product functions, to the best of our knowledge, this is the first time to study a fully end-to-end model with satisfiability analysis.

Deep neural network \cite{lecun2015deep,Goodfellow-et-al-2016-Book} has drawn attention in the past few years due to its impressive performance on NLP tasks.
Long Short-Term Memory (LSTM) \cite{hochreiter1997long} is shown to achieve the state-of-the-art results on many NLP tasks \cite{greff2015lstm,lample2016neural,tan2015lstm,nassif2016learning,wang2015long}. 
Attention mechanism \cite{NIPS2010_4089,denil2012learning} is effective in NLP tasks, such as machine translation\cite{bahdanau2014neural}, sentence summarization \cite{rush2015neural}, sentiment analysis \cite{tang-qin-liu:2016:EMNLP2016}, question and answering \cite{li2016dataset} and reading comprehension \cite{kumar2015ask,xiong2016dynamic}.
There are also studies of neural sequence labeling \cite{lample2016neural,ma2016end}.
However, traditional sequence labeling takes a single sequence as the to-be-labeled input. 
The proposed task naturally has two inputs: the question with the to-be-labeled tokens and the answers with the polarity.
Inspired by the task of reading comprehension, we take one more step to fuse the question and the answer together as a story and perform reading comprehension on such a QA story. 
Instead of learning question-aware representations of the story as in reading comprehension, we learn question-aware (or answer-aware) representations of the QA story as the side information to enrich the representation of the question (or the answer, resp.).

\section{Model Overview and Preliminary}
Formally, we define the 2 inputs sequence labeling problem as following.
Given a QA pair $(\mathbf{x}^q, \mathbf{x}^a)$ (we use bold symbols to indicate sequences),
we label each word in the question $\mathbf{x}^q$ as a label sequence $\mathbf{y}=(y_1, \dots, y_t, \dots, y_{T_q})$, where
$y_t\in L$ and $T_q$ is the length of the question.
Here $L$ is the label space and the two proposed tasks only differ in the label space $L$.
For product compatibility analysis, the label space is $L^c=\{\textit{O, C, I, U}\}$, indicating \textit{\underline{O}ther} non-entity words, \textit{\underline{C}ompatible}, \textit{\underline{I}ncompatible} and \textit{\underline{U}ncertain} entity words.
For product satisfiability analysis, the label space is $L^s=\{\textit{O, S, UN, U, F-S, F-UN, F-U}\}$, indicating \textit{\underline{O}ther} non-function words, \textit{\underline{S}atisfiable}, \textit{\underline{UN}satisfiabile}, \textit{\underline{U}ncertain} function target words, and \textit{\underline{S}atisfiable}, \textit{\underline{UN}satisfiabile}, \textit{\underline{U}ncertain} \textit{\underline{F}}unction words.

\begin{figure}[t]
    \centering    
    \includegraphics[width=3.2in]{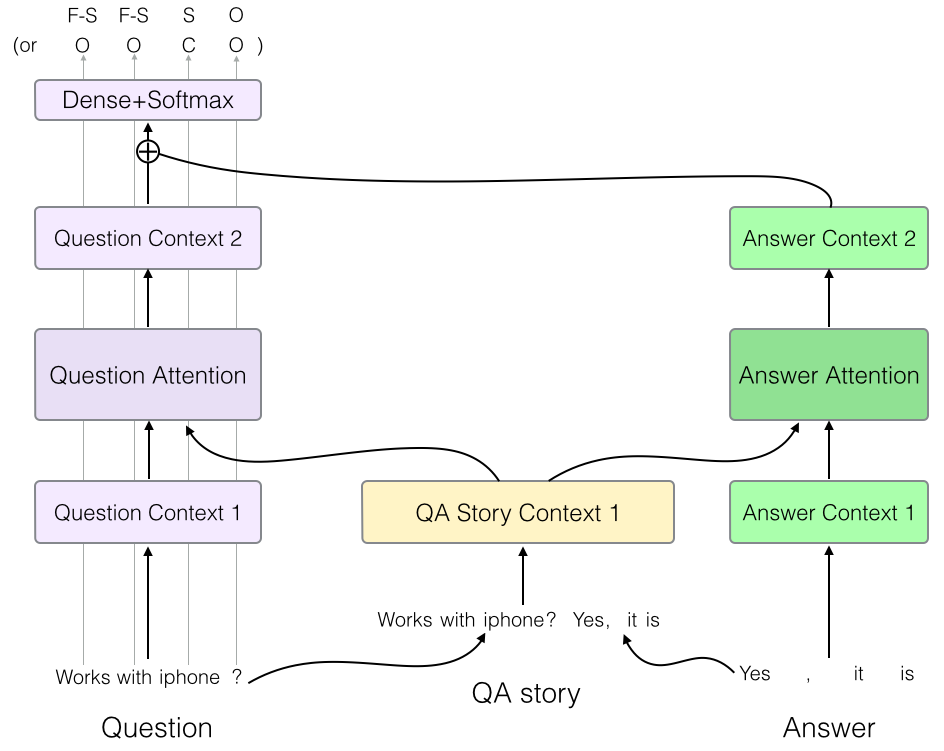}
        \caption{Dual Attention Network: given a question ``Works with iphone ?'' and its answer ``Yes, it is'', ``Works with iphone ?'' is can be labeled as $(\textit{F-S, F-S, S, O})$ for satisfiability analysis (or $(\textit{O, O, C, O})$ for compatibility analysis). ``Works with iphone'' is labeled as a satisfiable function expression, where ``iphone'' is a \underline{S}atifiable function targets (or \underline{C}ompatible entity) ) and ``Works with'' is labeled as satisfiable function words (\textit{F-S}). The details of question attention (and similarly for answer attention) is shown in Figure \ref{fig}.}
        \label{fig:fr}
    \end{figure}

The proposed network is illustrated in Figure~\ref{fig:fr}. 
The question and the answer are first concatenated to form a QA story $\mathbf{x}^\textit{qa}$.
Then the QA pair $(\mathbf{x}^q, \mathbf{x}^a)$ and the story $\mathbf{x}^\textit{qa}$ are passed into a shared embedding layer (not shown in the figure), followed by three respective BLSTM (Bidirectional Long Short-Term Memory \cite{hochreiter1997long,schuster1997bidirectional}) Context Layers to obtain contextual representations. 
So the vector representation at each position is encoded with the information from nearby words.
The contextual representation of the QA story is attended (read) by the contextual representations of the question and the answer, respectively.
This is done via two separate attention modules.
The attention process can be viewed as both the question and the answer read the QA story to form their corresponding side information.
Then the side information is concatenated with the original contextual representation for the question and the answer, respectively.
Now we call them QA-augmented question and answer. 
Later, we pass the QA-augmented question and answer representations to the Question Context 2 layer and the Answer Context 2 layer, respectively.
The second context layers here are used to learn the representations encoding both the original context representations and the QA story. 
Note that we only learn a single vector from the Answer Context 2 layer as the \emph{polarity vector} since we need a single vector to represent the polarity of the whole answer sequence.
Lastly, the polarity vector is duplicated $T_q$ times, each of which is concatenated to the representation of each word in the question, and output the label sequence $\mathbf{y}$ via a dense+softmax layer shared for each word in the question. 
Thus, both the question and the answer help to decide the output labels in an end-to-end manner. 

Note that more complicated deep architecture for sequence labeling can be leveraged (e.g., modeling label dependency using CRF or learning better features from character-level embedding as in LSTM-CRF \cite{lample2016neural} or LSTM-CNNs-CRF \cite{ma2016end}), and here we mainly focus on how to leverage a QA story to augment side information and perform sequence labeling. Next, we briefly introduce preliminary layers that are common in most NLP models.

\textbf{Input Layers}
Let the sequence $\mathbf{x}^q=(x_1^q, \dots, x_{T_q}^q )$ and $\mathbf{x}^a=(x_1^a, \dots, x_{T_a}^a )$ denote the question and the answer, respectively, where $T_a$ denote the length of the answer. 
A question (or an answer) may contain multiple sentences and we simply concatenate them into a single sequence.
We set $T_q=82$ as the maximum number of words in any question; since an answer can be as long as more than 2000 words, we simply make the answer the same length as the question ($T_a=82$) by removing words beyond the first 82 words.
Intuitively, the beginning of an answer is more informative.

We transform $\mathbf{x}^q$ ($\mathbf{x}^a$ and $\mathbf{x}^{qa}$, resp.) into an embedded representation $\boldsymbol{e}^q$ ($\boldsymbol{e}^a$ and $\boldsymbol{e}^{qa}$, resp.) via a word embedding matrix $W_e \in \mathbb{R}^{d_e \times V}$, and $d_e$ is the dimension (we set it as 300) of word vectors.
We pre-train the word embedding matrix $W_e$ via fasttext model \cite{bojanowski2016enriching} and fine tune the embeddings when optimizing the proposed model. 
The fasttext model allows us to obtain embeddings for out-of-vocabulary words (which is common in product QAs) from character n-grams embeddings.
The pre-training is discussed in Section Experimental Result. 

\textbf{BLSTM Context Layers}
The embedded word sequences ($\boldsymbol{e}^q$, $\boldsymbol{e}^a$ and $\boldsymbol{e}^\textit{qa}$) are fed into the Question Context 1 layer, the Answer Context 1 layer, and the QA Story Context layer, respectively. 
BLSTM \cite{hochreiter1997long,schuster1997bidirectional} is an important variant of RNN due to its ability to model long-term dependencies and contexts in both forward and backward directions in a sequence. 
The key component of an LSTM unit is the memory cell, which avoids overwriting the hidden state at every time step. 
An LSTM unit decides to update the memory cell via input, forget and output gates. 
We set all the output dimensions of BLSTM layers as 128.
We omit the details of the update mechanism and interested readers may refer to \cite{hochreiter1997long} for details. 
Note that other variants of RNN such as GRU \cite{chung2014empirical} can also be used, here we mainly focus on how attention mechanism can help improve the performance. 
After passing the question and the answer embedding through these BLSTM layers, we have the hidden representations $\boldsymbol{h}^{q, 1}$, $\boldsymbol{h}^{a, 1}$ and $\boldsymbol{h}^\textit{qa}$ for the question, answer and QA story, respectively.

\section{Dual Attention Network}
\label{sec:dan}
\subsection{Attention-based Module}
Next, we leverage attention mechanism to allow both the question and the answer to enrich their representations.
Attention mechanism \cite{NIPS2010_4089,denil2012learning} is popular in recent years due to its capability of modeling variable length memories rather than fixed-length memory. 
We utilize attention mechanism to synthesize side information from the QA story. 
We introduce two attention mechanisms: \emph{question attention} and \emph{answer attention}. 
By reading the QA story, they both get side information by the fact that the question and the answer in a QA pair are connected.
Intuitively, the words in the answer depend on the question and the question can help infer compatibility information from the answer. 
According to our experience of dataset annotation, we find that some entities are hard to label by reading only the question or the answer. 
However, if we read the question and answer as a whole, more often we get the idea of what the QA pair discusses about. 
For example, a question like ``straight talk?'' for a cell phone can be hard to label. 
If we have an answer ``yes, it works with straight talk well.'', we can somehow guess ``straight talk'' should be a carrier. 
Similarly, the ``straight talk'' indicated by the question also helps us to identify the polarity word ``well'' in the answer. 
This is very important in identifying implicit polarities in the answer.
We mimic this procedure of human's reading comprehension using the following attention mechanism.  

\begin{figure}
    \centering    
    \includegraphics[width=3.2in]{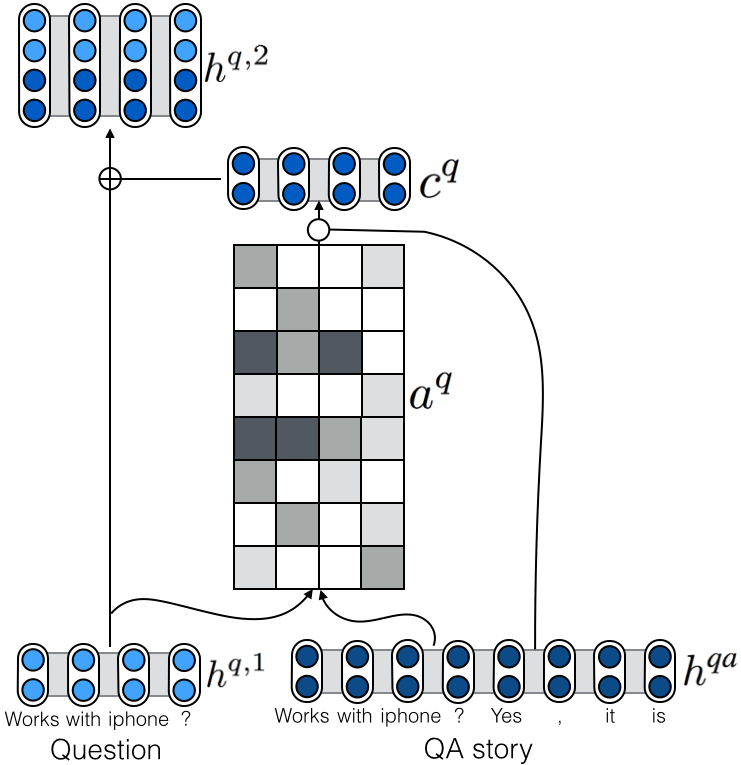}
    \caption{Question Attention: each word in the question attends on each word in the QA story.}
    \label{fig}
\end{figure} 

Let the output from the QA story Context layer be $\boldsymbol{h}^\textit{qa}=(h_1^\textit{qa}, \dots, h_{T_q}^\textit{qa}, h_{T_q+1}^\textit{qa}, \dots, h_{T_q+T_a}^\textit{qa} )$. 
For the question (answer) attention, we obtain the attention weight for the $t$-th question (answer) word when reading the $u$-th word in the QA story via a dot product and further normalized by a softmax function:
\begin{equation}
\begin{split}
a_{t, u}^q=\frac{\exp{\big((h_t^{q,1})^T h_u^{\textit{qa}}\big)} }{\sum_{u'=1}^{T_q+T_a} \exp{\big((h_t^{q,1})^T h_{u'}^{\textit{qa}}\big)} } .
\end{split}
\end{equation}
Then the contextual information for the $t$-th word in the question (or answer) is the weighted sum over all words in the QA story:
\begin{equation}
\begin{split}
c_t^q=\sum_{u=1}^{T_q+T_a} a_{t, u}^q h_u^{\textit{qa}}.
\end{split}
\end{equation}
We concatenate $c_t^q$ with $h_t^{q,1}$ as the representation of the $t$-th word in the question (answer): $h_t^{q,2}=h_t^{q,1} \oplus c_t^q$. 
Similarly, we have another answer attention module to obtain a sequence of answer word representations and the $t$-th word in the answer is denoted as $h_t^{a,2}=h_t^{a,1} \oplus c_t^a$. 

\subsection{Stacked Structures}
Next, $\bm{h}^{q,2}$ and $\bm{h}^{a,2}$ are fed into the Question Context 2 layer and the Answer Context 2 layer, respectively, which are similar to the stacked BLSTM \cite{el1995hierarchical} to obtain better representation of the sequences. 
We utilize two structures of BLSTMs: the many-to-many structure on questions and the many-to-one structure for learning the answer representation since we care about the answer polarity more than word-by-word representations. 
The answer representation is a concatenation of the last and the first outputs of a forward and backward LSTMs, respectively. Finally, we have $\boldsymbol{h}^{q,3}$ for the question sequence and $h^{a,3}$ for the polarity representation of the whole answer. 

\subsection{Joint Model}
Now we form the joint model in an end-to-end manner, by merging the question branch and the answer branch into prediction $\hat{\mathbf{y} }$. 
So the labels for each word in question can affect both the question and answer branches to learn better representations. 
To match the output length of the answer branch to the output length of the question branch, we obtain $T_q$ copies of $h^{a,3}$ and concatenate each copy with the output of the question branch at each word position. 
Then we reduce the dimension of each concatenated output to $|L|$ via a fully-connected layer by weights $W$ and bias $b$ shared among all positions of the question:

\begin{equation}
\begin{split}
s_t^q & =W[ h_{t}^{q,3} \oplus h^{a, 3} ]+b , \\
\end{split}
\end{equation}
where $s_t^q \in \mathbb{R}^{|L|}$ is the representation the $t$-th position in question. 
We output the probability distribution over label space $L$ for the $t$-th question word via a softmax function: 
\begin{equation}
\begin{split}
p^q(\hat{y}_t=l; \Theta) & =\frac{\exp(s_{t, l}^q)}{\sum_{k=1}^{\left\vert{L}\right\vert}\exp(s_{t, k}^q)} ,
\end{split}
\end{equation}
where $\Theta$ represents all trainable parameters.

Finally, we optimize the cross entropy loss function:
\begin{equation} \label{eq:opt}
\begin{split}
J(\Theta)=-\sum_m^{|M|} \sum_t^{|T_q|} \sum_l^{|L|} y_{t, l}^{(m)} \log p^q(\hat{y}_t^{(m)}=l; \Theta) ,
\end{split}
\end{equation}
where $M$ represents all the training examples and $y_{t, l}^{(m)} \in \{0, 1\}$ is the ground truth for the $t$-th question word and label $l$ in the $m$-th training example. So $y_t^{(m)}$ is a one-hot vector. 
We leverage Adam optimizer \cite{kingma2014adam} to optimize this loss function and set the learning rate as 0.001 and keep other parameters the same as the original paper. 
We set the dropout rate to 0.1. The batch size is set to 128.

During testing, the prediction for each position in the question is computed as:
\begin{equation} \label{eq:pred}
\begin{split}
\hat{y_t}=\operatorname*{arg\,max}_{l \in L} p^q(\hat{y}_t=l; \Theta) .
\end{split}
\end{equation}
Lastly, for function satisfiability analysis, we extract function words and function targets with polarities over label space $L^s$; 
for compatibility analysis, we extract complementary entities with polarities over label space $L^c$.

\section{Experimental Result}
\label{sec:exp}
In this section, we discuss the details of the annotated corpus, and experimentally demonstrate the superior performance of DAN.
\subsection{Corpus Annotation and Analysis}
\label{sec:data}

\begin{table}
\footnotesize
\centering
\scalebox{0.95}{
\begin{tabular}{ l | c | c | c | c  }
\hline
Product & QA & \% with Fun. & Intr. Fun. & Extr. Fun. \\
\hline
DSLR & 319 & 27.9 & 57 & 32 \\
E-Reader & 270 & 44.07 & 48 & 71 \\
Speaker & 153 & 38.56 & 14 & 45 \\
Tablet & 300 & 41.67 & 31 & 94 \\
Cellphone 1 & 168 & 60.12 & 6 & 95 \\
Cellphone 2 & 321 & 43.61 & 25 & 115 \\
Laptop 1 & 289 & 25.26 & 42 & 31 \\
Laptop 2 & 404 & 58.17 & 56 & 179 \\
Netbook & 194 & 46.91 & 15 & 76 \\
TV & 291 & 49.48 & 21 & 123 \\
TV Console & 167 & 50.9 & 7 & 78 \\
Gaming Console & 204 & 75.49 & 14 & 140 \\
Apple Watch & 324 & 37.65 & 55 & 67 \\
\hline
VR Headset & 438 & 76.94 & 12 & 325 \\
Stylus & 260 & 72.69 & 13 & 176 \\
Micro SD Card & 281 & 81.85 & 1 & 229 \\
Mouse & 254 & 74.41 & 30 & 159 \\
Tablet Stand & 212 & 88.68 & 3 & 185 \\
\hline 
\end{tabular}
}
\caption{Function statistics of 18 selected labeled products. QAs: number of QA pairs; \% with Fun.: percentage of QA pairs containing function needs; Intr. Fun.: QAs about intrinsic functions; Extr. Fun.: QAs about extrinsic functions.}
\label{table:dataset_fun}
\end{table}

We crawled about 1 million QA pairs from the web pages of products in the electronics department of Amazon.
These 1 million QAs combined with all electronics customer reviews \cite{McAPanLes15} are used to train word embeddings.
We use customer reviews because the texts in QAs are too short to train good quality embeddings.
The combined corpus is 4 GB.

We select 42 products with 7969 QA pairs in total as the to-be-annotated corpus.
The corpus is labeled by 3 annotators independently. The general annotation guidelines are as the following:
\begin{enumerate}
\item only yes/no QAs should be labeled;
\item a function expression should be labeled as either intrinsic function or extrinsic function;
\item each function expression is labeled with separate function words and function targets;
\item function words are verbs and prepositions around the function targets;
\item function targets are token spans containing nouns, adjectives, or model numbers; 
\item abstract entities such as ``picture'', ``video'', etc. are considered as function targets for intrinsic functions;
\item specific entities are considered as complementary entities (function targets for extrinsic functions). They are not limited to products from Amazon, but also include general entities like ``phone'', or service providers like ``AT\&T'';
\item implicit yes/no answers should also be labeled to increase the recall rate;
\item implicit answers without direct experience on the target product are labeled as uncertain answers (e.g., ``I am not sure but it works for my android phone.'').
\end{enumerate}
All annotators initially agreed on their annotations on 83\% of all QA pairs.
Disagreements are then resolved to reach final consensus annotations.

\begin{table}
\footnotesize
\centering
\scalebox{0.95}{
\begin{tabular}{ l | c | c | c | c }
\hline
Product & Satis. & Unsatis. & Uncertain & \#Desc. \\
\hline
DSLR & 62 & 9 & 18 & 7\\
E-Reader & 66 & 35 & 18 & 8\\
Speaker & 35 & 16 & 8 & 7\\
Tablet & 76 & 22 & 27 & 7\\
Cellphone 1 & 88 & 2 & 11 & 6\\
Cellphone 2 & 63 & 47 & 30 & 39\\
Laptop 1 & 59 & 3 & 11 & 4\\
Laptop 2 & 98 & 88 & 49 & 5\\
Netbook & 59 & 12 & 20 & 11\\
TV & 84 & 38 & 22 & 21\\
TV Console & 46 & 27 & 12 & 21\\
Gaming Console & 78 & 29 & 47 & 19\\
Apple Watch & 70 & 40 & 12 & 8\\
\hline
VR Headset & 106 & 200 & 31 & 8\\
Stylus & 97 & 24 & 68 & 13\\
Micro SD Card & 168 & 16 & 46 & 9\\
Mouse & 106 & 36 & 47 & 5\\
Tablet Stand & 95 & 30 & 63 & 55\\
\hline 
\end{tabular}
}
\caption{Statistics of 18 selected labeled products on satisfiability. Satis.: satisfiable function needs; Unsatis.: unsatisfiable function needs; Uncertain: uncertain function needs; \#Desc.: number of compatible products mentioned in the corresponding product descriptions.}
\label{table:dataset_s}
\end{table}

Due to limited space, the statistics of 18 selected products with annotations are shown in Table \ref{table:dataset_fun} and \ref{table:dataset_s}.
We can see that the majority functions are extrinsic functions, indicating the importance of product compatibility analysis.
This is close to our common sense as complementary entities can be unlimited, whereas the intrinsic functions are usually limited.
We observe that accessories (the last 5 products) have a higher percentage of 
functionality related questions than main products (the first 13 products).
This is as expected since accessories are poorly described in the product description and accessories usually have many complementary entities. 
From Table \ref{table:dataset_s}, we can see that
the polarity distribution is not even: most products have more satisfiable functions than unsatisfiable or uncertain ones.
This is because customers are more likely to ask a question to confirm functionality before purchasing, and many unsatisfiable functions are thus identified in advance without asking a question.
The only exception is the relatively new product \textit{VR headset}, which is problematic due to its short time of testing on the market.

We further investigate product descriptions of these 18 products and count the number of compatible products mentioned, as shown in the last column of Table \ref{table:dataset_s}.
Interestingly, no incompatible entities can be found, justifying the need for compatibility analysis on incompatible products from user-generated data.

The corpus is preprocessed using Stanford CoreNLP \footnote{http://stanfordnlp.github.io/CoreNLP/} regarding sentence segmentation, tokenization, POS-tagging, lemmatizing and dependency parsing. 
The last 3 steps provide features for the Conditional Random Fields (CRF) \cite{lafferty2001conditional} baseline. 
We shuffle all QA pairs and select 70\% of QA pairs for training, 10\% for validation and 20\% for testing.

\subsection{Baselines}
We compare DAN with the following baselines.\\
\textbf{CRF}: This baseline is to show that a traditional sequence labeling model performs poorly. 
Note that CRF \cite{lafferty2001conditional} can only be evaluated on extraction without polarity detection since it cannot incorporate the answer into the model. 
We train CRF models using Mallet\footnote{\url{http://mallet.cs.umass.edu/}}.
We use the following features: the words within a 5-word window, the POS tags within a 5-word window, the number of characters, binary indicators (camel case, digits, dashes, slashes and periods), and dependency relations for the current word obtained via dependency parsing.\\
\textbf{QA S-BLSTM}: This baseline does not have any attention module. We use it to show that attention mechanism improves the results.\\
\textbf{QA CoAttention}: This baseline is inspired by \cite{xiong2016dynamic}, where the question and the answer part directly attend to each other, without form the QA story. We use this baseline to demonstrate that attending on a QA story is better.\\
\textbf{DAN (-) Answer Attention}: This baseline does not have the answer attention module in DAN. We use this baseline to show that the answer attention also helps to improve the performance on polarity detection.

\subsection{Product Compatibility Analysis}

\begin{table}[t]
\small
\centering
\scalebox{1}{
\begin{tabular}{ l | c | c | c }
\hline
Method & PCA $\mathcal{F}_1$ & CER $\mathcal{F}_1$ & Polar. Acc. \\
\hline
CRF & - & 71.0 & -\\
\hline
QA S-BLSTM & 61.3 & 78.0 & 81.6 \\ 
\hline
QA CoAttention & 62.9 & 79.8 & 82.4 \\
\hline
DAN (-) Ans. Attention & 63.9 & 80.2 & 81.5 \\
\hline
DAN & \textbf{64.8} & \textbf{80.9} & \textbf{82.5} \\
\hline 
\end{tabular}
}
\caption{Performance on Product Compatibility Analysis (PCA): PCA $\mathcal{F}_1$ is the averaged $\mathcal{F}_1$ scores over 3 types (compatible, incompatible and uncertain) of complementary entities; CER $\mathcal{F}_1$ only evaluates the performance of \underline{C}omplementary \underline{E}ntity \underline{R}ecognition without considering the polarity; Polarity Accuracy only evalutes the polarity detection given successfully identified complementary entities. }
\label{table:pca}
\end{table}

We first evaluate the performance of product compatibility analysis.
Note that the label space of this task is $L^c=\{\textit{O, C, I, U}\}$, as described in Section Model Overview and Preliminary.
We consider an extracted entity that has more than or equal to 50\% overlapping words with the ground truth entity as a \textit{positive extraction}.
The polarity of a positive extraction is computed as the majority type voted from all words in such a positive extraction. 
So a \textit{true positive example} must have at least 50\% overlapping words and a match on polarity.
Any positive extraction with no corresponding ground truth entity is considered as a \textit{false positive example}.
Any example with mismatched polarity or negative extraction is considered as a \textit{false negative example}.
We average the $\mathcal{F}_1$ computed from the above definition as the PCA $\mathcal{F}_1$ column, shown in Table \ref{table:pca}.
Further, by only considering a positive extraction as a true positive and a negative extraction as a false negative, we compute CER $\mathcal{F}_1$ for complementary entity recognition.
Given a positive extraction, we further compute the classification accuracy over 3 polarity types to show the effectiveness of polarity detection.

\textbf{Result Analysis}: 
We can see that the performance of DAN outperforms other baselines on PCA $\mathcal{F}_1$, CER $\mathcal{F}_1$, and polarity accuracy.
The attention mechanism boosts the performance of CER a lot.
With the attention on the answer, the polarity is more accurate when DAN is compared with DAN (-) Answer Attention.
The baseline QA CoAttention indicates that attending on the QA story is better than attending the question or the answer alone.
Lastly, CRF performs poorly on learning better word representations.

\subsection{Product Function Satisfiability Analysis}

\begin{table}[t]
\small
\centering
\scalebox{1}{
\begin{tabular}{ l | c | c | c }
\hline
Method & FSA $\mathcal{F}_1$ & FNR $\mathcal{F}_1$ & Polar. Acc. \\
\hline
CRF & - & 70.2 & -\\
\hline
QA S-BLSTM & 61.1 & 77.0 & 82.2 \\ 
\hline
QA CoAttention & 61.7 & 77.4 & 81.3 \\
\hline
DAN (-) Ans. Attention & 62.5 & 78.0 & 83.0 \\
\hline
DAN & \textbf{63.9} & \textbf{78.2} & \textbf{84.3} \\
\hline 
\end{tabular}
}
\caption{Performance on product Function Satisfiability Analysis (FSA): FSA $\mathcal{F}_1$ is the averaged $\mathcal{F}_1$ scores over 3 types (satisfiable, unsatisfiable and uncertain) of extracted function expressions; FNR $\mathcal{F}_1$ only evaluates the performance of \underline{F}unction \underline{N}eed (function expression) \underline{R}ecognition without considering the polarity; Polarity Accuracy only evalutes the polarity detection given successfully identified function needs. }
\label{table:fsa}
\end{table}

We then evaluate the performance of product function satisfiability analysis, which requires the label space of all models to be $L^s=\{\textit{O, S, UN, U, F-S, F-UN, F-U}\}$.
Similar to the previous task, we consider an extracted function target that has more than or equal to 50\% overlapping words with the ground truth function target as a \textit{positive function target extraction}.
If there is at least one function words that are correctly predicted, or the function word from the ground truth is missing, we consider such case as a \textit{positive function word extraction}.
A \textit{true positive extraction} is generated when both a positive function target extraction and a positive function word extraction happen.
The rest of evaluation metrics is the same as in the previous subsection except the change of corresponding terms, as shown in Table \ref{table:fsa}.

\textbf{Result Analysis}: 
We can see that the performance of DAN outperforms other baselines. 
DAN improves over DAN (-) Answer Attention a little for function need recognition, but a lot for polarity detection due to answer attention.
The baseline QA CoAttention indicates that the QA story in DAN is better given the longer QA story rather than the question or the answer alone.
Further, we notice that the performance of QA S-BLSTM and QA CoAttention are close.
So the short question or answer alone may not have enough information. 
Sometimes it may even introduce noise. 
Lastly, CRF performs poorly because of its poor representation learning capability.

\subsection{Qualitative Analysis}
To get a better sense of the extracted function expressions (or needs), we sample a few predictions for 5 popular products from DAN, as shown in Table \ref{table:qua}. 
We observe that many function needs are indeed customers' high priority needs.
Most function needs are extrinsic functions and their function targets can be interpreted as complementary products.
For example, it is important to know that the Tablet is not designed for high-performance games like ``fallout 4'', 
or google apps are not runnable on Cellphone 1.
Intrinsic functions are also identified, such as ``waterproof'' or ``support multi pairing''.
Knowing whether Apple Watch is waterproof or not is very important when deciding whether to buy such a product.

\begin{table}
\footnotesize
\centering
\scalebox{0.85}{
\begin{tabular}{ l | l | l }
\hline
Product & Satisfiable & Unsatisfiable \\
\hline
Tablet & 
\begin{tabular}[c]{@{}l@{}} 
\textbf{run} \underline{itunes} \\
\textbf{read} \underline{kindle books} \\
\end{tabular}
& 
\begin{tabular}[c]{@{}l@{}} 
\textbf{play} \underline{Xbox One games} \\
\textbf{run} \underline{fallout 4} \\
\end{tabular}
\\
\hline
Cellphone 1 & 
\begin{tabular}[c]{@{}l@{}} 
\textbf{work on} \underline{cricket} \\
\textbf{run} \underline{Gmail} \\
\end{tabular}
& 
\begin{tabular}[c]{@{}l@{}} 
\textbf{get} \underline{google apps} \\
\textbf{support} \underline{SD card} \\
\end{tabular}
\\
\hline
Laptop 1 & 
\begin{tabular}[c]{@{}l@{}} 
\textbf{Install} \underline{Windows} \\
\textbf{install} \underline{adobe flash player} \\
\end{tabular}
& 
\begin{tabular}[c]{@{}l@{}} 
\textbf{use with} \underline{HDMI}\\
\textbf{support} \underline{Appleworks} \\
\end{tabular}
\\
\hline
Apple Watch & 
\begin{tabular}[c]{@{}l@{}} 
\textbf{work on} \underline{IPhone 6 plus} \\
\textbf{comp. with} \underline{apple comp.} \\
\end{tabular}
& 
\begin{tabular}[c]{@{}l@{}} 
\textbf{comp. with} \underline{android phones} \\
\underline{waterproof} \\
\end{tabular} \\
\hline
Mouse & 
\begin{tabular}[c]{@{}l@{}} 
\underline{mac} \textbf{compatible} \\
\underline{Android} \\
\end{tabular}
& 
\begin{tabular}[c]{@{}l@{}} 
\textbf{work on} \underline{glass table} \\
\textbf{support} \underline{multi pairing} \\
\end{tabular}
\\
\hline 
\end{tabular}
}
\caption{Satisfiable function needs and unsatisfiable function needs from 5 popular products: the function word are bolded and the function targets are underlined.}
\label{table:qua}
\end{table}

\section{Conclusion}
In this paper, we propose two closely related problems:
product compatibility analysis and function satisfiability analysis, where the second problem is a generalization of the first problem. 
We address this problem by first creating an annotated corpus based on Product Community Question and Answering (PCQA). 
Then we propose a neural Dual Attention Network (DAN) to solve these two (2) problems in an end-to-end manner. 
Experiments demonstrate that DAN is superior to a wide spectrum of baselines. 
Applications of this model can be found in e-commerce websites and recommender systems.

\section{ Acknowledgments}
This work is supported in part by NSF through grants IIS-1526499, CNS-1626432, and NSFC 61672313. 
We would also like to thank anonymous reviewers for their valuable feedback to improve this paper.

\bibliography{aaai2018}
\bibliographystyle{aaai}

\end{document}